\documentclass{article}

\usepackage{microtype}
\usepackage{graphicx}
\usepackage{subfigure}
\usepackage{booktabs} 
\usepackage{multirow} 
\usepackage{multicol} 
\usepackage{mathtools}
\usepackage{xspace}
\usepackage{amsmath}
\usepackage{amsfonts}

\usepackage{hyperref}

\DeclareMathOperator{\softplus}{softplus}
\DeclareMathOperator{\Be}{Be}
\DeclareMathOperator{\ELBO}{ELBO}

\DeclarePairedDelimiterX{\infdivx}[2]{(}{)}{%
  #1\;\delimsize\|\;#2%
}
\newcommand{\dkl}[2]{\ensuremath{KL\infdivx{#1}{#2}}\xspace}

\usepackage[accepted]{icml2021}

\icmltitlerunning{Transformation Models for Flexible Posteriors in Variational Bayes}

\begin{document}

\twocolumn[

\icmltitle{Transformation Models for Flexible Posteriors in Variational Bayes}




\begin{icmlauthorlist}
\icmlauthor{Sefan H{\"o}rtling}{htwg}
\icmlauthor{Daniel Dold}{htwg}
\icmlauthor{Oliver D{\"u}rr}{htwg}
\icmlauthor{Beate Sick}{zhaw_uzh}
\end{icmlauthorlist}

\icmlaffiliation{zhaw_uzh}{IDP, Zurich University of Applied Sciences, Switzerland, and EBPI, University of Zurich, Switzerland}
\icmlaffiliation{htwg}{IOS, Konstanz University of Applied Sciences, Germany}

\icmlcorrespondingauthor{Stefan  H{\"o}rtling}{stefan.hoertling@htwg-konstanz.de}
\icmlcorrespondingauthor{Oliver Dürr}{oliver.duerr@htwg-konstanz.de}
\icmlcorrespondingauthor{Beate Sick}{sick@zhaw.ch}

\icmlkeywords{Deep Neural Networks, Bayesian Neural Networks, Variational Inference, Normalizing Flows, Bernstein Polynomials}

\vskip 0.3in
]



\printAffiliationsAndNotice{}  

\begin{abstract}

The main challenge in Bayesian models is to determine the posterior for the model parameters. 
Already, in models with only one or few parameters, the analytical posterior can only be determined in special settings. In Bayesian neural networks, variational inference is widely used to approximate difficult-to-compute posteriors by variational distributions. Usually, Gaussians are used as variational distributions (Gaussian-VI) which limits the quality of the approximation due to their limited flexibility. Transformation models on the other hand are flexible enough to fit any distribution. Here we present transformation model-based variational inference (TM-VI) and demonstrate that it allows to accurately approximate complex posteriors in models with one parameter and also works in a mean-field fashion for multi-parameter models like neural networks.

\end{abstract}

\section{Introduction}
\label{sec:intro}

Uncertainty quantification is important, especially if model predictions are used to support high stake decision-making. Quantifying uncertainty in statistical or machine learning models is often achieved by Bayesian approaches, where the uncertainty of the estimated model parameters are represented by posterior distributions. Determining these posterior distributions analytically is usually impossible if the posterior takes a complex shape or if the model has many parameters, such as a neural network (NN). Variational inference (VI) is a well established method to approximate difficult-to-compute distributions through optimization \cite{jordan1999, wainwright2008, bleiVI}. In VI the posterior is approximated by a variational distribution by minimizing the Kullback-Leibler divergence between the variational distribution and the posterior. Usually, this is done by first choosing a family of parametric distributions, usually Gaussians, and then tuning the parameters of the variational distribution until its distance to the posterior is minimized. Obviously, the quality of the VI approximation depends on the similarity of the true posterior with the optimized member of the chosen distribution family. In cases where the posterior takes a complex shape, a simple variational distribution, such as a Gaussian, can never yield a good approximation. 

We propose to use transformation models (TMs) \cite{hothorn2014} to approximate complex posteriors via VI. The main advantage of TMs is their guarantee that any distribution shape can be achieved without predefining the family of the distribution. In this paper, we show how to combine TMs and VI to accurately approximate flexible posteriors for all parameters in variational Bayes models via a computational efficient optimization process. Moreover, we benchmark our approach against exact Bayesian models, MCMC-Simulations, and Gaussian-VI approximations.

 \section{Related Work}
\label{sec:related_work}

TMs were developed in the statistics community and have the focus on modeling a potentially complex conditional outcome distributions in regression models \cite{hothorn2014}. The main idea of TMs is to learn a potentially complex transformation function that transforms a simple base distribution, such as the Standard Gaussian $N(0,1)$, to a potentially complex outcome distribution under which the likelihood of the observed outcomes is maximized. The choice of the simple base distribution is unimportant for prediction purposes but gets crucial for inference \cite{hothorn2018}. Up to now, TMs were used to model unconditional distributions or conditional outcome distributions in statistical or deep learning regression models \cite{kook_herzog2020, buri2020b, sick2021, baumann2020}. Recently, a first Bayesian version of TMs were proposed \cite{carlan2020bayesian}, which yields exact posteriors by Hamilton Monte Carlo sampling, but is restricted to relatively small models and requires experience with designing priors. 

Independently to TMs, normalizing flows were developed in the deep learning community \cite{dinh2014} and are based on the same idea as TMs. Normalizing flows learn a chain of many simple, bijective transformations and are mainly used to model unconditional high-dimensional distributions for generative models \cite{kobyzev2020NFreview}. In generative deep learning models, VI was used to approximate the posterior of the latent variables \cite{rezende2015} and, indirectly, by constructing a flexible mixing density, to build the variational distribution of the weights from multivariate Gaussians~\cite{louizos2017multiplicative}. 

To the best of our knowledge, TM based or normalizing flow based VI that directly approximates the posteriors of all model parameters, such as the weights in a NN model, were not yet developed.

 When using VI for models with many parameters, such as NNs, it is usually not possible to optimize a joint variational distribution over all parameters that accounts for all potential dependencies between the parameters. In such cases, mean-field VI is used where the variational distributions of the parameters are optimized independently from each other. For deeper NNs it has been demonstrated that Gaussian-VI achieves the same quality in terms of the modeled outcome distribution regardless if the posteriors were optimized independently from each other or if correlations between the parameters were taken into account \cite{farquhar2020}.


\section{Methods}
\label{sec:methods}

In the following, we describe our proposed TM-VI approach, where we use transformation models in variational inference to achieve accurate approximations to potentially complex posteriors for parameters in Bayesian models. The code is publicly available on github\footnote{\url{https://github.com/stefan1893/TM-VI}}. The main idea is to enable the VI procedure to approximate the posterior of the model parameters by a flexible variational distribution. In TMs a complex target distribution of interest is fitted by learning a bijective transformation function $h$ that transforms between latent variable $z$ following a fixed simple distribution, e.g. $z \sim N(0,1)$, and a variable of interest following a potentially complex target distribution (see Figure \ref{fig:flow}). Our target distribution of interest is the variational distribution that approximates the posterior of model parameters (e.g. the weights $w$ of a NN).

\subsection{Transformation model} \label{sec:tm}
\begin{figure}[!h]
    \centering
    \includegraphics[width=0.45\textwidth]
    {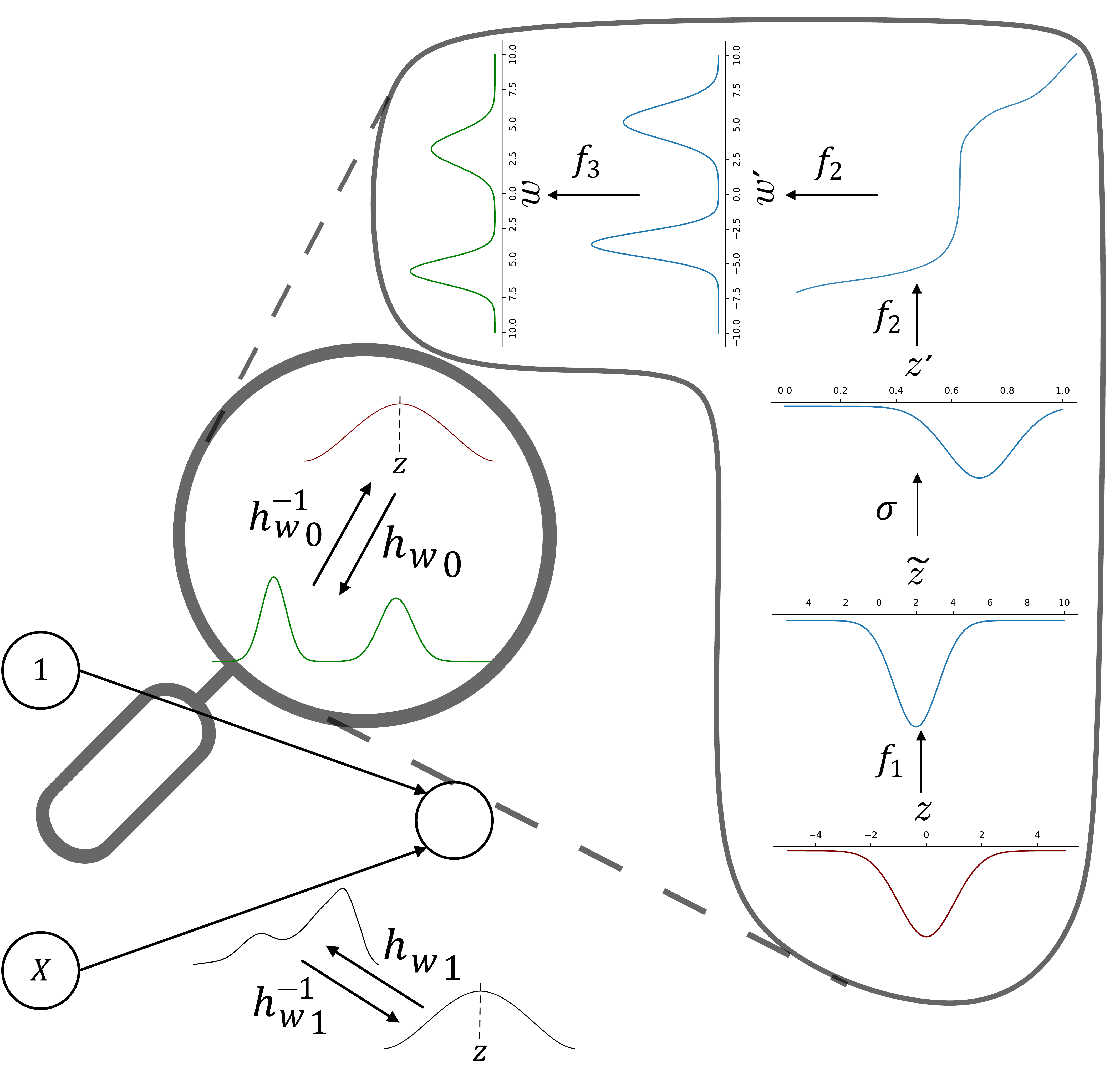}
    \caption{Overview of the transformation $h$ for modeling a potentially complex variational distribution $q_\lambda(w)$ of a  unconstrained weight $w$ in a Bayesian NN (lower left). The transformation $h$ is a chain of transformations, $ h=f_3 \circ f_2 \circ \sigma \circ f_1 $, that starts from a simple distribution, here $N(0,1)$ (depicted in red), and ends with the complex distribution $q_{\lambda}(w)$ (depicted in green). The first part of the flow, $\sigma \circ f_1$, transforms $N(0,1)$ into a distribution with support $[0,1]$, the flexible Bernstein polynomial used in $f_2$ allow for the creation of a complex shaped distribution, and $f_3$ yields the variational distribution $q_{\lambda}(w)$.}
    \label{fig:flow}
\end{figure}

The complete transformation function $h(z)=w$  consisting of a chain of  transformations $ h=f_3 \circ f_2 \circ \sigma \circ f_1 $ as shown in Figure \ref{fig:flow}. To achieve a bijective overall transformation $h$, it is sufficient that each transformation $f_i$ is a strictly monotone increasing transformation. A first scale and shift transformation $f_1(z) = a \cdot z + b$ followed by a sigmoid function transforms the standard Normal distributed $z$ into $[0,1]$. To ensure a strictly monotonic increase of $f_1$, we constrain the slope $a$ to be positive. 

The core of the transformation is the flexible Bernstein polynomial  of degree $M$:
\begin{equation}
\small
\label{eq:mlt}
    f_2(z') = \sum_{i=0}^M {\Be_i}(z') \frac{\vartheta_i}{M+1}
\end{equation}
With ${\Be_i}(z')$ being densities of Beta-functions which are defined on $z' \in [0,1]$. It is known that the Bernstein polynomials can uniformly approximate every function in $z \in [0,1]$~\cite{bernvstein1912demonstration}, see \cite{farouki2012bernstein} for a further discussion. An additional benefit of the Bernstein polynomials is that a strict a monotonic increase of $f_2(z')$ w.r.t. $z'$ can be achieved by simply enforcing that the Bernstein coefficients  are increasing, i.e.  $\vartheta_0 < \vartheta_1 < \ldots < \vartheta_M$. 
The last transformation is again a scale and shift transformation $f_3(w') = \alpha \cdot w' + \beta$ for which we constrain $\alpha$ to be positive to ensure a monotone increasing transformation. Altogether the complete transformation $h(z)$ is described by $M+5$ variational parameters $\lambda=(a,b,\vartheta_0, \ldots \vartheta_{M}, \alpha, \beta)$. 

We apply the following manipulations to the unrestricted parameters $\lambda'=(a',b,\vartheta_0', \ldots \vartheta_{M}', \alpha', \beta)$ to ensure the above described constrains of the parameters that guarantee a bijective transformation: $a=\softplus(a')$, $\alpha=\softplus(\alpha')$, $\vartheta_0=\vartheta'_0$, and $\vartheta_i = \vartheta_{i-1} + \softplus(\vartheta'_{i})$ for $i=1,\dots,M$.

To facilitate the fitting of distributions with potentially complex shapes or long tails, we use the fact that $f_2(0)=\vartheta_0$ and $f_2(1)=\vartheta_M$ (see \cite{ramasinghe2021robust}) to initialize our weights such that the Bernstein transformation $f_2$ yields a distribution which assigns substantial probability mass over the support of $w'$ before optimization. This can be achieved by defining a range on $w'$ with $w'_{min}$ and $w'_{max}$ and initializing $\vartheta'$ with $\vartheta'_0= w'_{min}$ and $\vartheta'_i = \softplus^{-1}((w'_{max} - w'_{min})/M)$ for $i=1,\dots,M$. Thus, they have the same initial support regardless of the degree of the Bernstein polynomial.

\subsection{Transformation model based variational inference} \label{sec:vi}
In VI the variational parameters $\lambda=(a,b,\vartheta_0, \ldots \vartheta_{M}, \alpha, \beta)$ are tuned such that the resulting variational distribution $q_\lambda(w)$ is as close to the posterior $p(w|D)$ as possible. This is done by minimizing the $KL$ divergence between the variational distribution and the (unknown) posterior:
\begin{multline}
\small
\dkl{q_\lambda(w)}{p(w|D)} = 
\int~q_\lambda(w)\log\left(\frac{q_\lambda(w)}{p(w|D)}\right)dw \\=
\log(D)  
- \underbrace{
\left( \mathbb{E}_{w \sim q_\lambda} (\log(p(D|w))) - \dkl{q_\lambda(w)}{p(w)}  \right)
}_{\ELBO(\lambda)}
\label{eq:vi1}    
\end{multline}
Instead of minimizing equation (\ref{eq:vi1}) usually only the 
evidence lower bound ELBO (see e.g. \cite{blundell2015weight}) is maximized which consists of the expected value of the log-likelihood, $\mathbb{E}_{w \sim q_\lambda} (\log(p(D|w)))$, minus the $KL$ divergence between the variational distribution $q_\lambda(w)$ and the known prior $p(w)$.
In practice, the negative ELBO is minimized by gradient descent. We approximate the expected log-likelihood by averaging over $T$ weight samples $w_t \sim q_\lambda(w)$. To get these weight samples we first draw $T$ samples $z_t \sim N(0,1)$ and then compute the corresponding weight samples via $w_t = h(z_t)$. We can approximate the expected log-likelihood for the training data by: 
\begin{equation}
\small
\mathbb{E}_{w \sim q_\lambda} (\log(p(D|w))) \approx \frac{1}{T} \sum_{t,i} \log\left(p(D_i|w_t)\right)
\label{eq:vi}    
\end{equation}

We use the same weight samples $w_t$ to approximate the Kullback-Leibler divergence between the variational distribution and the prior via:
\begin{equation}
\small
\dkl{q_\lambda(w)}{p(w)} \approx 
 \frac{1}{T} \sum_{t} \log\left(\frac{q_\lambda(w_t)}{p(w_t)}\right)
\label{eq:vi_sample_KL}    
\end{equation}
where the probability density $q_\lambda(w_t)$ can be calculated, from the samples $z_t$ using the change of variable function as:

\begin{equation}
\small
q_\lambda(w_t) = p(z_t) \cdot \left| \frac{\partial h_\lambda(z_t)}{\partial z} \right|^{-1}
\label{eq:vi_change}    
\end{equation}

\section{Results and discussion}
\label{sec:results}

We performed a couple of experiments to benchmark our TM-VI approach versus exact Bayesian solutions and Gaussian-VI.

\subsection{Models with a single parameter} 

We first discuss two experiments with models containing only one parameter. This has the advantage that we can rule out the mean field assumption as a potential reason for observed deficits of the achieved variational distribution.

\begin{figure}[h]
    \centering
    \includegraphics[width=0.45\textwidth]
    {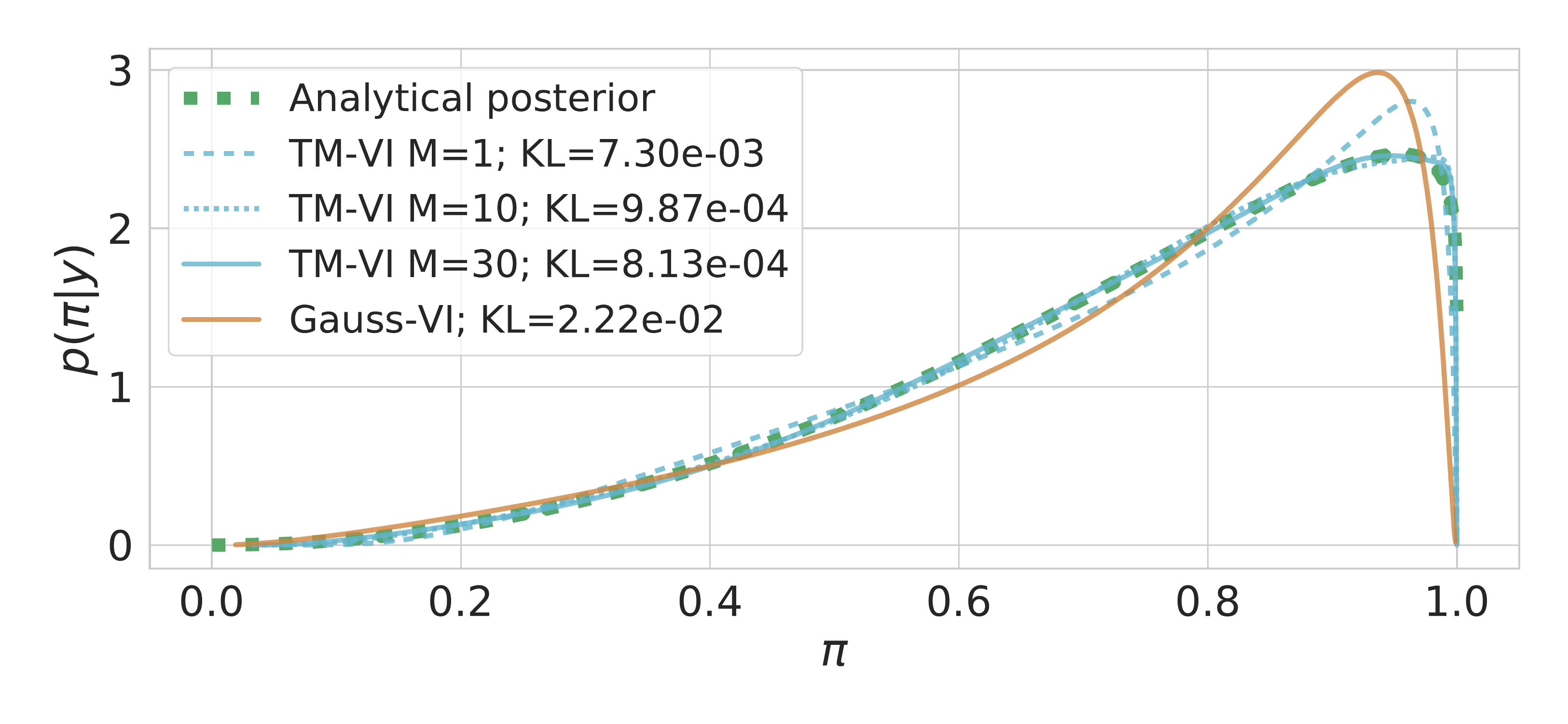}
    \caption{Comparison of the analytical posterior for the parameter $\pi$ in the Bernoulli model ${y \sim Ber(\pi)}$ with variational distributions achieved via TM-VI and Gaussian-VI. The blue lines show the results from our TM-VI model with different degrees M of the Bernstein polynomial shown in different line styles. In addition the $KL$ divergence between the variational distributions and the analytical posterior $\dkl{q_\lambda(w)}{p(w|D)}$ is shown in the legend. For an animated version, showing the training process for TM-VI with $M=10$ see: \url{https://youtu.be/_RA7QirjXMM}}
    \label{fig:bernoulli}
\end{figure}

\paragraph{Bernoulli example}
\ \\
We first look at an unconditional Bayesian model for a random variable $y$ following a Bernoulli distribution ${y \sim Ber(\pi)}$ which we fit based on only two samples ($y_1=1$, $y_2=1$). 
In this simple Bernoulli model, it is possible to determine the Bayesian solution analytically. Since the parameter $\pi$ can only take values between zero and one we choose a Beta-distribution $p(\pi)=Beta(\alpha=1.1,\beta=1.1)$ as prior which leads to the conjugated posterior $p(\pi|\text{data}) =Beta(\alpha + \sum{y_i}, \beta+n-\sum{y_i})$ (see analytical posterior in Figure \ref{fig:bernoulli}). 

\begin{figure}[!h]
    \centering
    \includegraphics[width=0.45\textwidth]
    {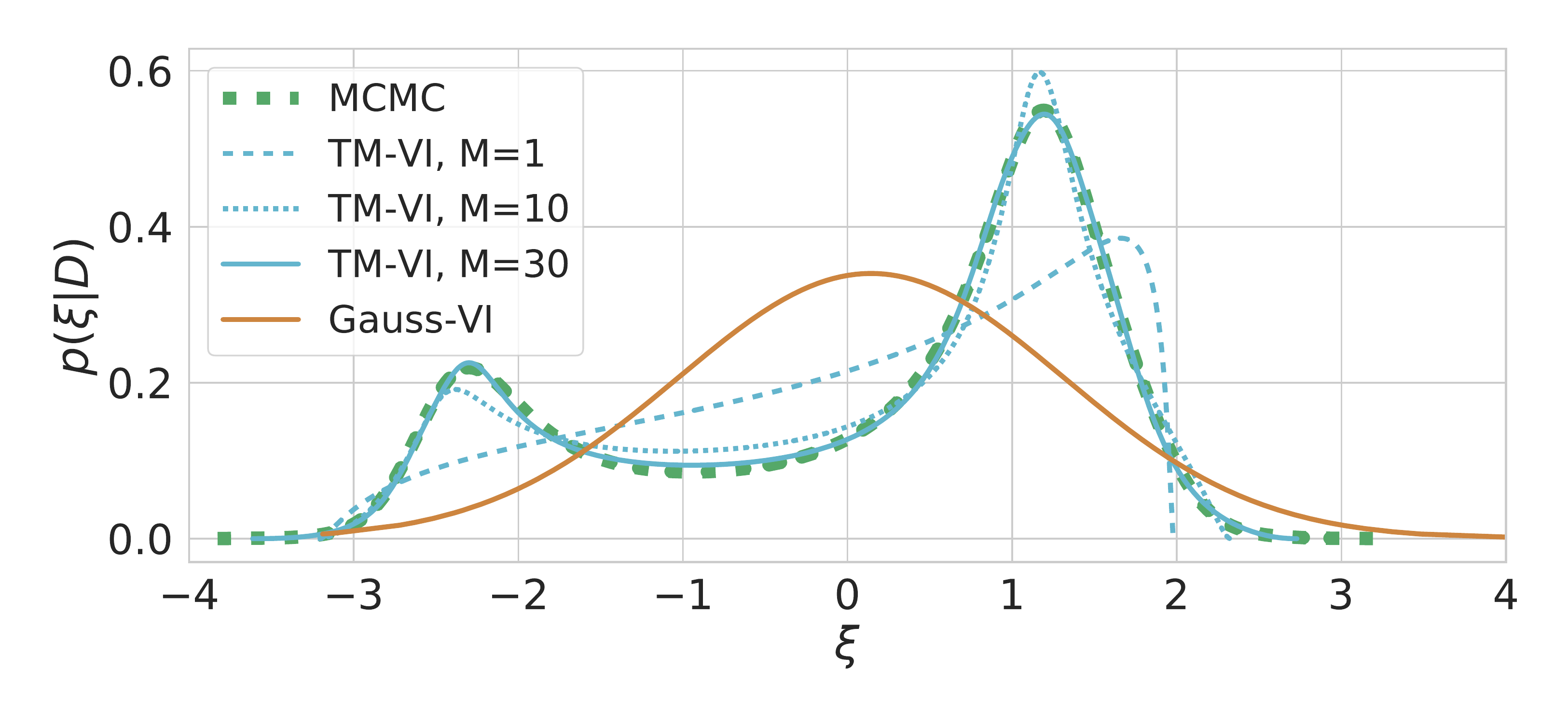}
    \caption{Posterior distribution of the parameter $\xi$ in the miss-specified Cauchy model $y\sim \text{Cauchy}(\xi, \gamma$). Comparing the bi-modal true posterior resulting from MCMC with variational distributions estimated via Gaussian-VI or TM-VI shows that the flexibility of TM-VI is needed to capture the bi-modal shape of the posterior.}
    \label{fig:cauchy}
\end{figure}

We now use our TM-VI method to approximate the posterior. To investigate how the flexibility of the Bernstein polynomial impacts the quality of the achieved variational distribution, we have used Bernstein polynomials with different degree $M$. To enforce the modeled variational distribution to be restricted on $[0,1]$ we pipe the result of $f_3$ (see Figure \ref{fig:flow}) through a sigmoid transformation to get the variational distribution of the parameter $\pi$. Figure \ref{fig:bernoulli} shows the achieved variational distributions after minimizing the negative ELBO via gradient descent as described in section \ref{sec:vi}. With increasing degree $M$ of the Bernstein polynomial, the variational distributions gets closer to the posterior (see $\dkl{q_\lambda(w)}{p(w|D)}$ in the legend of Figure \ref{fig:bernoulli}). Using the TM-VI with $M=30$ yields a variation distribution that approximates the posterior very accurately.  As expected, the Gaussian based VI has not enough flexibility to lead to a variational distribution that approximates the analytical posterior nicely (see Figure \ref{fig:bernoulli}).
\paragraph{Cauchy example}
\ \\
For the next experiment we follow an example from \cite{yao2020stacking} and we fit an unconditional Cauchy model $y \sim \text{Cauchy}(\xi, \gamma$) to six samples which we have drawn from mixture-Cauchy distribution $y \sim \text{Cauchy}(\xi_1=-2.5, \gamma)+\text{Cauchy}(\xi_2=2.5, \gamma)$. Because of the miss-specification of the model, the true posterior of the parameter $\xi$ has a bi-modal shape which we have determined via MCMC (see Figure~\ref{fig:cauchy}).  
We have used TM-VI and Gaussian-VI to approximate the posterior of the Cauchy parameter $\xi$ by a variational distribution. Because the possible range of $\xi$ is not restricted we can use the result of $f_3$ (see Figure \ref{fig:flow}) as variational distribution for $\xi$. While TM-VI has enough flexibility to capture the bi-modal shape of the posterior, Gauss-VI fails as expected.

\subsection{Multi-weight neural networks} 
\begin{figure}[!h]
    \centering
    \includegraphics[width=0.45\textwidth]
    {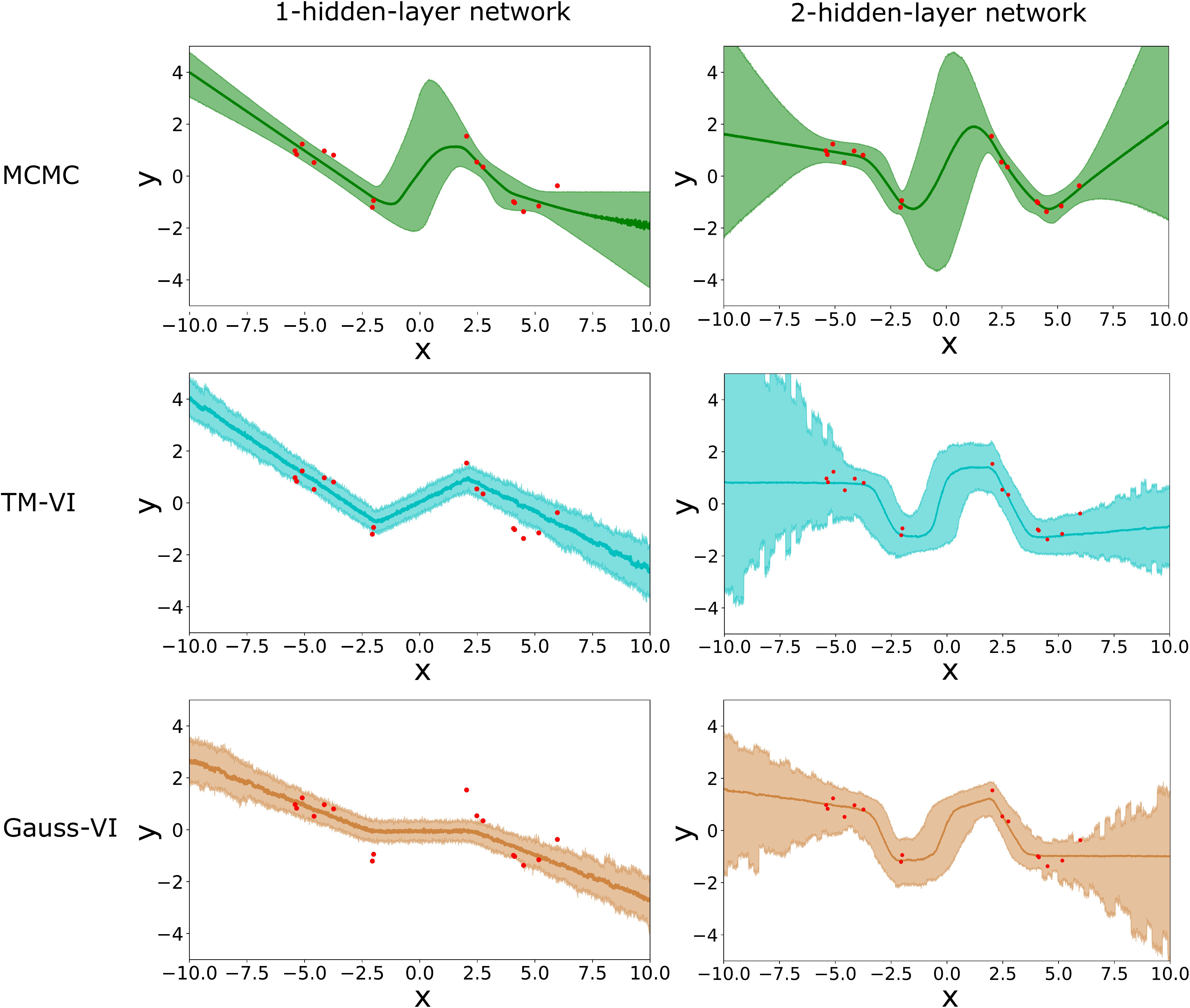}
    \caption{Posterior distribution for the conditional mean $\mu(x)$  of the conditional outcome distribution $N(\mu(x),\sigma)$ modeled by a Bayesian NN with 1 hidden layer (left panel) or 2 hidden layers (right panel). The Bayesian solution was determined via MCMC (first row) or via mean-field TM-VI (second row) or via mean-field Gaussian-VI (third row).}
    \label{fig:nn}
\end{figure}

We now investigate how our TM-VI approach performs in conditional multi-parameter models like NNs.
For this experiment, we sample 16 data points clustered in two regions of a noisy sinus wave (see points in Figure \ref{fig:nn}). We then use  MCMC, TM-VI, and Gaussian-VI to determine the solution of a Bayesian NN  which controls the conditional mean $\mu(x)$ of the conditional outcome distribution $N(\mu(x),\sigma)$. We use two different NNs: one with 1 hidden layer and 3 neurons per layer model, and one with 2 hidden layers and 10 neurons per layer. 
The variational distributions for the weights propagate to the distribution of the conditional mean $\mu(x)$.
Figure \ref{fig:nn} demonstrates, that in the smaller 1-hidden-layer NN, mean-field TM-VI slightly outperforms mean-field Gaussian-VI. In the larger 2-hidden-layer NN, both mean-field VI approaches perform comparably. Both, mean-field Gaussian-VI and TM-VI, fail to capture the uncertainty in-between the two clusters of data points.  We attribute both observations to the used mean-field approach, which is known to underestimate the uncertainty \cite{bleiVI} and probably also masks the benefit of the flexible TM-VI.   

\section{Conclusion and outlook}
\label{sec:outlook}
We have introduced TM-VI to achieve flexible variational distributions that accurately approximate potential complex parameter posteriors. In single-parameter models with a complex posterior, we have shown that TM-VI perfectly approximates the posterior while Gaussian-VI fails. In multi-parameter NN models, we have demonstrated that TM-VI can be used in a mean-field fashion. For small NNs, our TM-VI approach produces slightly superior results compared to the Gaussian-VI, but the limitations of the mean-field approach are clearly visible. In the future, we plan to extend TM-VI for Bayesian inference in models with few interpretable parameters by dropping the mean-field assumption to get accurate posterior approximations. 

\section{Acknowledgements}
\label{sec:acknowledgements}
Part of the work has been founded by the Federal	Ministry of	Education	and	Research	of	Germany	(BMBF) in the project DeepDoubt (grant no. 01IS19083A).

\bibliography{main}

\begin{thebibliography}{20}
\providecommand{\natexlab}[1]{#1}
\providecommand{\url}[1]{\texttt{#1}}
\expandafter\ifx\csname urlstyle\endcsname\relax
  \providecommand{\doi}[1]{doi: #1}\else
  \providecommand{\doi}{doi: \begingroup \urlstyle{rm}\Url}\fi

\bibitem[Baumann et~al.(2020)Baumann, Hothorn, and R{\"u}gamer]{baumann2020}
Baumann, P.~F., Hothorn, T., and R{\"u}gamer, D.
\newblock Deep conditional transformation models.
\newblock \emph{arXiv preprint arXiv:2010.07860}, 2020.

\bibitem[Bern{\v{s}}te{\i}n(1912)]{bernvstein1912demonstration}
Bern{\v{s}}te{\i}n, S.
\newblock D{\'e}monstration du th{\'e}oreme de weierstrass fond{\'e}e sur le
  calcul des probabilities.
\newblock \emph{Comm. Soc. Math. Kharkov}, 13:\penalty0 1--2, 1912.

\bibitem[Blei et~al.(2017)Blei, Kucukelbir, and McAuliffe]{bleiVI}
Blei, D.~M., Kucukelbir, A., and McAuliffe, J.~D.
\newblock Variational inference: A review for statisticians.
\newblock \emph{Journal of the American statistical Association}, 112\penalty0
  (518):\penalty0 859--877, 2017.

\bibitem[Blundell et~al.(2015)Blundell, Cornebise, Kavukcuoglu, and
  Wierstra]{blundell2015weight}
Blundell, C., Cornebise, J., Kavukcuoglu, K., and Wierstra, D.
\newblock Weight uncertainty in neural network.
\newblock In \emph{International Conference on Machine Learning}, pp.\
  1613--1622. PMLR, 2015.

\bibitem[Buri et~al.(2020)Buri, Curt, Steeves, and Hothorn]{buri2020b}
Buri, M., Curt, A., Steeves, J., and Hothorn, T.
\newblock Baseline-adjusted proportional odds models for the quantification of
  treatment effects in trials with ordinal sum score outcomes.
\newblock \emph{BMC medical research methodology}, 20:\penalty0 1--14, 2020.

\bibitem[Carlan et~al.(2020)Carlan, Kneib, and Klein]{carlan2020bayesian}
Carlan, M., Kneib, T., and Klein, N.
\newblock Bayesian conditional transformation models.
\newblock \emph{arXiv preprint arXiv:2012.11016}, 2020.

\bibitem[Dinh et~al.(2014)Dinh, Krueger, and Bengio]{dinh2014}
Dinh, L., Krueger, D., and Bengio, Y.
\newblock Nice: Non-linear independent components estimation.
\newblock \emph{arXiv preprint arXiv:1410.8516}, 2014.

\bibitem[Farouki(2012)]{farouki2012bernstein}
Farouki, R.~T.
\newblock The bernstein polynomial basis: A centennial retrospective.
\newblock \emph{Computer Aided Geometric Design}, 29\penalty0 (6):\penalty0
  379--419, 2012.

\bibitem[Farquhar et~al.(2020)Farquhar, Smith, and Gal]{farquhar2020}
Farquhar, S., Smith, L., and Gal, Y.
\newblock Liberty or depth: Deep bayesian neural nets do not need complex
  weight posterior approximations.
\newblock \emph{arXiv e-prints}, pp.\  arXiv--2002, 2020.

\bibitem[Hothorn et~al.(2014)Hothorn, Kneib, and B{\"u}hlmann]{hothorn2014}
Hothorn, T., Kneib, T., and B{\"u}hlmann, P.
\newblock Conditional transformation models.
\newblock \emph{Journal of the Royal Statistical Society: Series B: Statistical
  Methodology}, pp.\  3--27, 2014.

\bibitem[Hothorn et~al.(2018)Hothorn, Moest, and Buehlmann]{hothorn2018}
Hothorn, T., Moest, L., and Buehlmann, P.
\newblock Most likely transformations.
\newblock \emph{Scandinavian Journal of Statistics}, 45\penalty0 (1):\penalty0
  110--134, 2018.

\bibitem[Jordan et~al.(1999)Jordan, Ghahramani, Jaakkola, and Saul]{jordan1999}
Jordan, M.~I., Ghahramani, Z., Jaakkola, T.~S., and Saul, L.~K.
\newblock An introduction to variational methods for graphical models.
\newblock \emph{Machine learning}, 37\penalty0 (2):\penalty0 183--233, 1999.

\bibitem[Kobyzev et~al.(2020)Kobyzev, Prince, and
  Brubaker]{kobyzev2020NFreview}
Kobyzev, I., Prince, S., and Brubaker, M.
\newblock Normalizing flows: An introduction and review of current methods.
\newblock \emph{IEEE Transactions on Pattern Analysis and Machine
  Intelligence}, 2020.

\bibitem[Kook et~al.(2020)Kook, Herzog, Hothorn, D{\"u}rr, and
  Sick]{kook_herzog2020}
Kook, L., Herzog, L., Hothorn, T., D{\"u}rr, O., and Sick, B.
\newblock Deep and interpretable regression models for ordinal outcomes.
\newblock \emph{arXiv preprint arXiv:2010.08376}, 2020.

\bibitem[Louizos \& Welling(2017)Louizos and
  Welling]{louizos2017multiplicative}
Louizos, C. and Welling, M.
\newblock Multiplicative normalizing flows for variational bayesian neural
  networks.
\newblock In \emph{International Conference on Machine Learning}, pp.\
  2218--2227. PMLR, 2017.

\bibitem[Ramasinghe et~al.(2021)Ramasinghe, Fernando, Khan, and
  Barnes]{ramasinghe2021robust}
Ramasinghe, S., Fernando, K., Khan, S., and Barnes, N.
\newblock Robust normalizing flows using bernstein-type polynomials.
\newblock \emph{arXiv preprint arXiv:2102.03509}, 2021.

\bibitem[Rezende \& Mohamed(2015)Rezende and Mohamed]{rezende2015}
Rezende, D. and Mohamed, S.
\newblock Variational inference with normalizing flows.
\newblock In \emph{International Conference on Machine Learning}, pp.\
  1530--1538. PMLR, 2015.

\bibitem[Sick et~al.(2021)Sick, Hathorn, and D{\"u}rr]{sick2021}
Sick, B., Hathorn, T., and D{\"u}rr, O.
\newblock Deep transformation models: Tackling complex regression problems with
  neural network based transformation models.
\newblock In \emph{2020 25th International Conference on Pattern Recognition
  (ICPR)}, pp.\  2476--2481. IEEE, 2021.

\bibitem[Wainwright \& Jordan(2008)Wainwright and Jordan]{wainwright2008}
Wainwright, M.~J. and Jordan, M.~I.
\newblock \emph{Graphical models, exponential families, and variational
  inference}.
\newblock Now Publishers Inc, 2008.

\bibitem[Yao et~al.(2020)Yao, Vehtari, and Gelman]{yao2020stacking}
Yao, Y., Vehtari, A., and Gelman, A.
\newblock Stacking for non-mixing bayesian computations: The curse and blessing
  of multimodal posteriors.
\newblock \emph{arXiv preprint arXiv:2006.12335}, 2020.

\end{thebibliography}
\bibliographystyle{icml2021}


\end{document}